\documentclass[runningheads,a4paper,orivec]{llncs}

\usepackage{times}
\usepackage{amssymb}
\usepackage{graphicx}
\usepackage{url}
\usepackage{multirow}
\usepackage[T1]{fontenc}
\usepackage{hyperref}
\usepackage{booktabs}
\usepackage{tabularx}
\usepackage{array}

\usepackage[T1]{fontenc}
\usepackage{tipa}

\usepackage{amsmath}
\newcommand{\keywords}[1]{\par\addvspace\baselineskip
\noindent\keywordname\enspace\ignorespaces#1}

\urldef{\mailsa}\path|johannes.wirth.3@iisys.de|

\begin{document}

\title{OLaPh: Optimal Language Phonemizer}

\titlerunning{OLaPh: Optimal Language Phonemizer}

\author{Johannes Wirth\inst{1}\orcidID{0009-0002-0666-7693}}

\authorrunning{Johannes Wirth}

\institute{Institute for Information Systems at Hof University of Applied Sciences, Alfons-Goppel-Platz 1, 95028 Hof, Germany \\
\url{https://www.iisys.de/} \\
\mailsa\\
}

\index{Wirth, Johannes}

\toctitle{} \tocauthor{}

\maketitle

%
%
%
%
\begin{abstract}
Phonemization is a critical component in text-to-speech synthesis. Traditional approaches rely on deterministic transformations and lexica, while neural methods offer potential for higher generalization on out-of-vocabulary (OOV) terms. We introduce OLaPh (Optimal Language Phonemizer), a hybrid framework that integrates extensive multilingual lexica with advanced NLP techniques and a statistical subword segmentation function. Evaluations on the WikiPron benchmark show OLaPh significantly outperforms established baselines in overall accuracy and maintains robustness on OOV data through advanced fallback mechanisms. To further explore neural generalization, we utilize the framework to synthesize a high-consistency training corpus for an instruction-tuned Large Language Model (LLM). While the deterministic framework remains more accurate overall, the LLM demonstrates strong generalization, matching or partly exceeding the framework's performance. This suggests that the LLM successfully internalized phonetic intuitions from the synthetic data that transcend the framework's capabilities. Together, these tools provide a comprehensive, open-source resource for multilingual grapheme-to-phoneme conversion (G2P) research.
\end{abstract}

\section{Introduction}
Phonemization or grapheme-to-phoneme (G2P) conversion is a core component of text-to-speech (TTS) systems, ensuring correct pronunciation, prosody, and intelligibility. Some modern TTS models rely solely on deep neural networks to infer pronunciation from raw text \cite{gao_e3_2023,yang24l_interspeech}, but this sacrifices control and adaptability. Users cannot manually correct rare words, loanwords, or domain-specific terms, and improving performance often requires resource-intensive retraining \cite{chen_f5-tts_2024}, which is impractical in many applications.

To retain flexibility and explainability, many TTS systems incorporate phonemization as an intermediate preprocessing step \cite{chen_f5-tts_2024,ren_fastspeech_2022}. Widely used approaches rely on rule-based transformations and lexicon lookups, which struggle with morphological complexity \cite{demberg_phonological_2007}, number normalization, named entities, and foreign words \cite{bisani_joint-sequence_2008}. Even more advanced frameworks that add preprocessing steps such as part-of-speech (POS) tagging and neural models for unseen words still produce systematic errors by treating out-of-vocabulary (OOV) words as purely statistical guesses, failing to achieve flawless results and limiting synthesized speech quality in real-world applications.

This work introduces OLaPh (Optimal Language Phonemizer), a framework designed to bridge the gap between deterministic precision and neural flexibility. OLaPh enhances traditional G2P workflows by leveraging expansive phonological lexica, advanced NLP integration with Named Entity Recognition (NER) and POS tagging, and a robust statistical subword segmentation function. Our evaluations across four languages demonstrate that OLaPh significantly reduces phoneme error rates compared to established tools and models, achieving high accuracy even on OOV inputs. Furthermore, we show that OLaPh can serve as a high-consistency data generator to train neural models: an instruction-tuned LLM trained on OLaPh-generated data shows strong generalization capabilities, in some cases resolving phonetic ambiguities that surpass the deterministic components of the framework.
\section{Related work}
Early research on phonemization established that a simple one-to-one alignment of graphemes  to phonemes is insufficient, even if this trivial approach correctly handles many cases, as most general rules in pronunciation have exceptions \cite[pp.~105-106]{dutoit_introduction_1997}. This led to two main approaches: One creates a dictionary of root morphemes, generating word derivations with modifiers and handling unknown words through rules. The other derives phonetic rules from a dictionary, applies them with priority, and stores only rule exceptions for lookup \cite{lammens_lexicon-based_1987}.

Modern phonemization systems can rely on much larger dictionaries and rulesets due to far less rigorous  storage limitations and the widespread availability of phoneme representations on public platforms like Wiktionary. A notable example is the phonemization component of the TTS engine eSpeakNG \cite{noauthor_espeak-ngespeak-ng_nodate} (eSpeak Next Generation), the successor to eSpeak \cite{noauthor_espeak_nodate}, which was previously used in commercial TTS systems like Google Translate \cite{noauthor_giving_nodate}. eSpeakNG follows the earlier discussed phonemization method of rule-based phonemization with exception lookup but features an extensive and well curated rule set for each of the 127 supported languages and accents, enabling accurate phonemization of many number structures, symbols, and complex contexts. However, the system is not able to distinguish most homographs and often phonemizes foreign-language words incorrectly, as shown in the following evaluations. 

Another popular phonemization system is Gruut \cite{noauthor_rhasspygruut_2025}, which utilizes  comparatively large lexica built from open-source data. Unlike eSpeakNG, it uses G2P models based on conditional random fields (CRFs) to predict pronunciations if a word is not found in the reference lexicon. Each of the 13 supported languages is associated with its own CRF-trained G2P model. Additionally, Gruut incorporates various parsers for number and symbol normalization and, in few languages, CRF-based POS tagging to correctly resolve phonological ambiguities, such as homographs (e.g., \textit{wound} /wu\textipa{\textlengthmark}nd/ as a noun versus \textit{wound} /wa\textipa{\textupsilon}nd/ as the past tense of the verb \textit{wind}). These additional components can be advantageous in edge cases compared to eSpeakNG, but they more frequently reduce the overall quality of phonemization, which is demonstrated in the evaluations that follow.

More recent  G2P approaches utilize deep neural networks (DNNs) to predict phoneme representations from text. T5G2P \cite{rezackova_t5g2p_2021} is a transformer-based T5 model \cite{raffel_exploring_2020} trained for English and Czech. Additionally, larger-scale models based on ByT5 \cite{xue_byt5_2022-1} have been developed for phonemization across 99 languages \cite{zhu22_interspeech}. ByT5, unlike token-based models such as the base T5 architecture, processes byte input and delivers improved performance in multilingual settings, as reported by the authors. ByT5 faces the same challenges as eSpeakNG and Gruut, as it was trained on lexicon data and data preprocessed by eSpeakNG, making it equally prone to errors in complex or context-dependent phonemization. Further research also investigates the use of large language models for phonemization \cite{chen_f5-tts_2024,qharabagh_llm-powered_2024}. However, direct performance comparisons between the reported quality of the mentioned models are challenging due to differences in evaluation data and metrics used. A general drawback of such DNN-based G2P approaches is their limited accuracy in languages without one-to-one grapheme-to-phoneme mappings and languages, where large datasets are scarce \cite{cheng_survey_2024}. 
\section{Approach}
To improve the capabilities of applied phonemization, OLaPh integrates larger lexica, NER, improved number normalization, POS tagging, language detection, and a novel statistical subword segmentation approach to address the shortcomings of the previously mentioned phonemization systems.
\subsection{Components}
\subsubsection{Lexicon lookup}\label{lexiconlookup}
OLaPh builds on Gruut's lexicon-based approach, using recent Wiktionary dumps to extract IPA transcriptions in four languages. While the system was evaluated on and currently supports the languages English, German, French and Spanish,  lexica for Danish, Dutch, Finnish, Italian, Swedish and Polish were also extracted for future integration and better coverage of loanwords. Additional lexica were created for character mapping, abbreviations, and symbols for every supported language. As Wiktionary pages are updated regulary, the lexica can be easily extended, improving direct lookup accuracy over time.
\subsubsection{Named Entity Recognition (NER)}
Proper names and organizations often follow non-standard or foreign pronunciation. Detecting entities prior to phonemization allows applying language-specific rules (e.g., “FBI” in German remains English). SpaCy models \cite{honnibal2020spacy} are used for NER in each language. Detected entities are phonemized via lexicon lookup using the identified source language if available. If a detected named entity cannot be phonemized in the source language, the word is skipped and processed through the following mechanisms.
\subsubsection{Number normalization}
Many phonemization systems struggle with numerical expressions, mathematical symbols, and special characters. By normalizing numbers (e.g., dates, times, decimals, ordinals) and symbols into their spoken equivalents before phonemization, inconsistencies in synthesized speech can be reduced. For the English language, the library num2words \cite{noauthor_savoirfairelinuxnum2words_2025} is used, which is also utilized in the Gruut framework. For German, a dedicated normalization algorithm was implemented, as its complex case system required a more refined solution.
\subsubsection{POS tagging}
Homographs such as the English word read (present tense /\textipa{\textturnr i\textlengthmark d}/ versus past tense /\textipa{\textturnr \textepsilon d}/), require context for accurate phonemization. The previously created lexica contain POS tags for homographs to help resolve such ambiguities by ensuring that pronunciation aligns with the correct usage. For POS tagging, the same spaCy models are used as for NER. This method also supports dialect phonemization if the lexicon contains corresponding word variants.
\subsubsection{Language detection}
Loanwords and code-switching \cite{zhou_end--end_2020} present challenges for phonemization, as different languages follow distinct pronunciation rules. By identifying the language of individual words or phrases (including named entities), language-specific lookups and operations can be performed, rather than enforcing a single-language phonemization, which would induce errors in such cases. For this task, the Lingua language detection framework \cite{stahl_pemistahllingua-rs_2025} was utilized as it supports a large number of languages and overall achieved usable  detection accuracies in experiments.
\subsubsection{Statistical Subword Segmentation}
When a word is not found in the extensive lexica, a statistical compound-splitting algorithm is applied. This algorithm segments the word into plausible subword candidates and scores each combination by aggregating the corpus frequencies of individual subwords, weighted by their relative lengths and a penalty for the total number of segments.  Subword frequencies are derived from a reference corpus of over 15 million sentences from a recent Wikipedia dump, processed for each supported language. In languages such as German, characterized by productive compounding and interleaved loanwords, linear left-to-right or right-to-left lexicon lookups often fail to identify optimal split points. To address this, the algorithm exhaustively evaluates all valid candidate segmentations and assigns a composite score to each based on the following principle:

\[
S\left( W \right)=\left(\sum_{s\in W}^{}F\left( s \right)\times \left( \frac{\left| s \right|}{\left| W \right|} \right)^{\alpha} \times L\left( s \right)\right) \times n^{-\beta},
\]
\\
where $W$ denotes a candidate segmentation (a combination of subwords) and $s$ a subword within that combination; 
 $S(W)$ the segmentation score of a subword combination.; 
 $F(s)$ The raw frequency (number of occurrences) of subword $s$ in the reference corpus; 
$\left(\tfrac{|s|}{|W|}\right)^{\alpha}$ the relative length of $s$ compared to the full word $W$, weighted by $\alpha$; 
$L(s)$ a length-based penalty ($0.1$ for single-letter subwords, $0.5$ for two letters, and $1$ otherwise); 
$n$ is the number of subwords in the combination, and $n^{-\beta}$ is a penalty favoring fewer segments, tunable by $\beta$ (default $15$).

Using recursive word segmentation, frequency-based scoring, and subword language detection, the algorithm identifies the optimal segmentation without relying on manual rules or language-specific grammars. This approach enables the phonemization framework to decompose complex compounds dynamically, significantly improving accuracy on structurally complex words across different linguistic families.
\subsection{Framework overview}
The components described in the previous subsection are integrated into a complete system organized in a hierarchical order. The process flow of the system is outlined below.

\begin{figure*}
    \centering
    \includegraphics[
    width=0.7\textwidth]{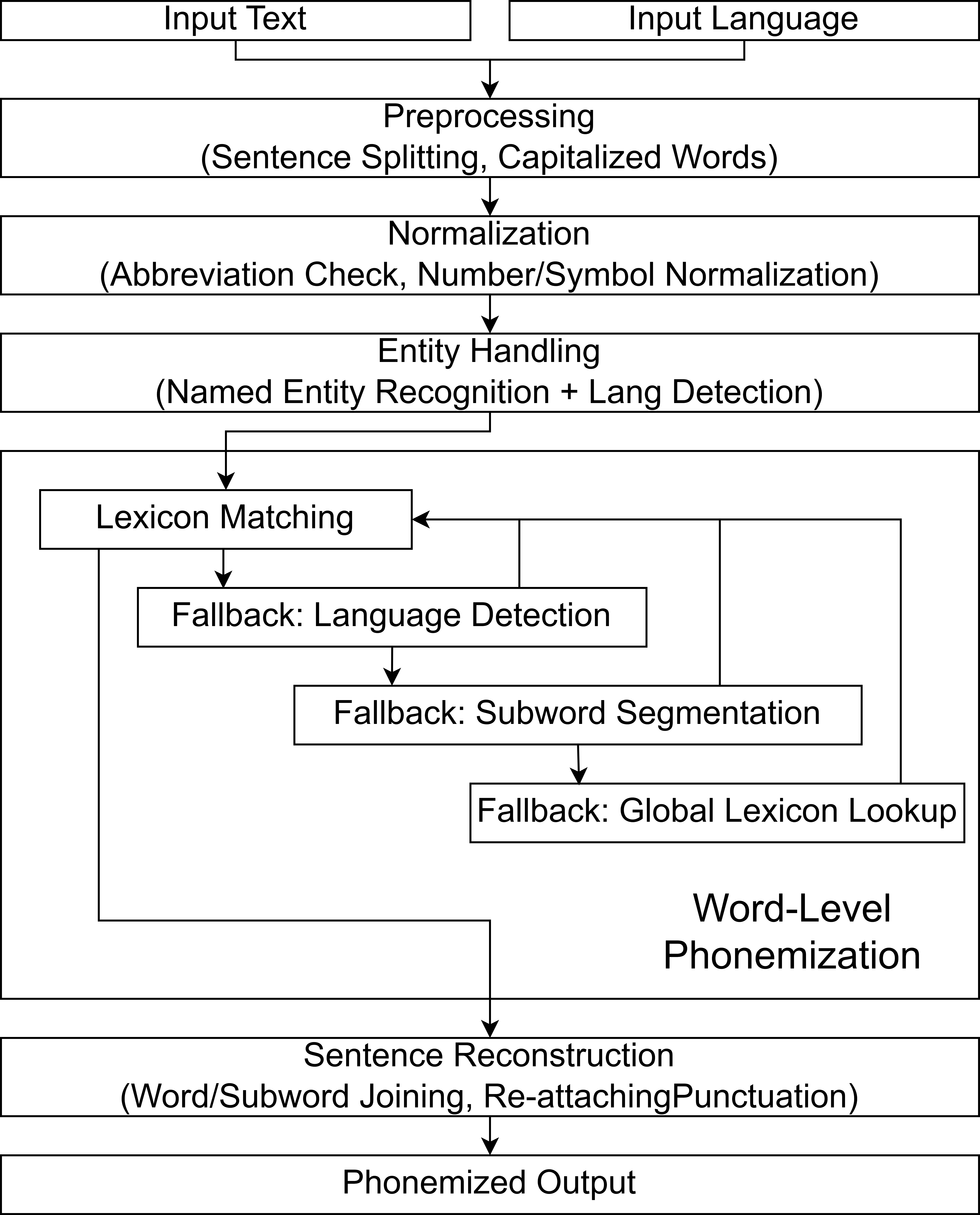}
    \caption{Workflow representation of the proposed phonemization framework OLaPh.}
    \label{fig:flowdiagram}
\end{figure*}

As shown in \autoref{fig:flowdiagram}, the workflow follows a sequential refinement architecture. First, the input is pre-processed for POS tagging, NER, and abbreviation resolution. Then, a multi-stage, word-level phonemization sequence is triggered: the system initially attempts a primary lexicon lookup, followed by a foreign-language check and subsequent lexicon lookup for the detected language. If the word remains unresolved, the system utilizes the statistical subword segmentation algorithm for complex compound decomposition in the detected language. As a final fallback, the system performs a global search across all available lexica; this handles cases where words remain unclassified or contain foreign symbols (e.g., Hanzi) not supported by the primary language's phonemic rules. Once all words and subwords are phonemized, the system performs a final reconstruction pass. This stage re-integrates the phonemic strings with the original punctuation, ensuring the output maintains the structural integrity of the input text.

\subsection{Extension: LLM Training with OLaPh-Generated Data}
Beyond the rule-based framework, OLaPh also serves as a high-fidelity data generator for finetuning neural phonemization models. To evaluate this utility, we finetuned a pretrained 2 billion parameter GemmaX-based LLM\cite{cui2025multilingualmachinetranslationopen} on grapheme-phoneme pairs synthesized by OLaPh from the FineWeb corpus\cite{penedo_fineweb_2024,penedo_fineweb2_2024}. The training set comprised 1,500,000 sentence pairs per target language (English, French, German, and Spanish), supplemented by 200,000 isolated, randomly selected lexicon entries per language, totaling 3 million training examples. To facilitate phonemic representation, the model's tokenizer was extended with 256 specialized phoneme tokens derived from the training data where 192 already preexisted in the model's tokenizer.
Unlike the OLaPh framework and other pipelines that rely on multiple components like spaCy for preprocessing, the resulting LLM learns to perform end-to-end G2P directly from the raw graphemic input. The model was instruction-fine-tuned using a supervised sequence-to-sequence objective. Each training sample was formatted with a natural language prompt: "translate this from [source language] to phonemes: [input graphemes]". This allows the model to treat phonemization as a translation task, an objective the GemmaX model was initially optimized for, leveraging its built-in linguistic knowledge to map text to IPA representations. The model was fine-tuned for a single epoch using a learning rate of $2 \times 10^{-5}$ with an inverse square root schedule. Training was conducted on two NVIDIA RTX PRO 6000 Blackwell GPUs, requiring  ~17 hours for a single epoch run. We utilized the AdamW optimizer with a weight decay of $0.01$ to ensure stable optimization of the extended phoneme vocabulary.
\section{Evaluations}
To evaluate the performance of OLaPh, we implemented the framework in Python and conducted a word-level comparative analysis against two established frameworks, eSpeak-NG and Gruut, as well as a ByT5 model fine-tuned for multilingual G2P \cite{zhu22_interspeech} trained on data generated by eSpeak-NG. Additionally, the previously described instruction-tuned GemmaX model (OLaPh LLM) was included to assess the performance of larger scale neural end-to-end approaches for this task. OLaPh framework and OLaPh LLM were further directly compared regarding their performance on out-of-vocabulary (OOV) data. 
\subsection{Benchmark Evaluation}
For the evaluation benchmark, we utilized the multilingual pronunciation data mined by WikiPron \cite{lee-etal-2020-massively}. This dataset provides broad coverage across all four target languages (English, French, German, and Spanish). Crucially, the benchmark includes terms not present in the primary OLaPh lexica, providing a OOV test set that was used to specifically evaluate the efficacy of the language detection and statistical subword segmentation functionality. Performances were measured using Phoneme Error Rate (PER) to capture the edit distance between predicted and reference IPA strings. The average PER reported in all tables is calculated as a weighted average across all samples from the four target languages.

\begin{table}[ht]
\centering
\caption{Wikipron benchmark Statistics with number of words covered by OLaPh lexica and OOV words.}
\label{tab:dataset_statistics}
\begin{tabularx}{0.8\textwidth}{@{} l *{5}{>{\centering\arraybackslash}X} @{}}
\toprule
\textbf{Language} & \textbf{In Vocabulary Words} & \textbf{OOV Words} & \textbf{Total Words} & \textbf{OOV Words (\%)} \\ \midrule
EN      & 62,989 & 4,730 & 67,719 & 6.98\\
FR          & 58,195  & 13,028 & 71,223  & 18.29\\
DE    & 37,903  & 4,970 & 42,873  & 11.59\\
ES    & 68,480  & 30,318 & 98,798  & 30.69\\ \bottomrule
\end{tabularx}
\end{table}

\autoref{tab:dataset_statistics} provides an overview of the grapheme-phoneme pairs per language taken from WikiPron provided datasets. For English, American English was selected as the Gruut framework specifically only supports this dialect while all other models support both. The columns for OOV words are listed in regard to how many samples can statically be looked up in lexica by OLaPh and how many need to be resolved through its advanced components. To ensure a fair comparison, benchmark labels were normalized by removing glottal stops and diacritics. These phonetic modifiers are broadly supported by OLaPh, but only partially or inconsistently implemented across the other evaluation frameworks (e.g., where OLaPh and eSpeak utilize explicit stress markers, gruut often introduces glottal stops). Stripping these non-segmental features establishes a consistent baseline across all models.

\begin{table}[ht]
\centering
\caption{Phonemization performance across all samples (average PER in percent).}
\label{tab:overall_results}
\begin{tabularx}{0.8\textwidth}{@{} l *{5}{>{\centering\arraybackslash}X} @{}}
\toprule
\textbf{Model} & \textbf{EN} & \textbf{FR} & \textbf{DE} & \textbf{ES} & \textbf{AVG} \\ \midrule
eSpeak-NG      & 14.59 & 6.20 & 17.59 & 4.32 & 9.84 \\
Gruut          & 16.55  & 4.05 & 17.56  & 4.21 & 10.29 \\
ByT5 small    & 14.81  & 7.55 & 26.16  & 5.43 & 9.91 \\ \midrule
\textbf{OLaPh (Ours)} & \textbf{10.10} & \textbf{2.17} & \textbf{3.77} & \textbf{2.58} & \textbf{5.02} \\
OLaPh LLM    & 13.88  & 7.55 & 11.42  & 3.18 & 8.69 \\ \bottomrule
\end{tabularx}
\end{table}

\autoref{tab:overall_results} presents a comparison of all frameworks on the full benchmark dataset. The results indicate that OLaPh is significantly more accurate than the next-best candidates. While the ByT5-based model achieved the worst results except for English and a tie with OLaph LLM in French, OLaPh LLM outperformed both Gruut and eSpeak-NG in English, German and Spanish.

\begin{table}[ht]
\centering
\caption{Phonemization performance across OLaPh OOV samples only (average PER in percent).}
\label{tab:oov_results}
\begin{tabularx}{0.8\textwidth}{@{} l *{5}{>{\centering\arraybackslash}X} @{}}
\toprule
\textbf{Model} & \textbf{EN} & \textbf{FR} & \textbf{DE} & \textbf{ES} & \textbf{AVG} \\ \midrule
eSpeak-NG      & 25.71 & 8.43 & 19.33 & 4.24 & 10.05 \\
Gruut          & 30.84  & 5.51 & 22.58  & 4.49 & 10.87 \\
ByT5 small    & 22.43  & 7.21 & 28.68  & 5.68 & 9.88 \\ \midrule
\textbf{OLaPh (Ours)} & \textbf{16.61} & \textbf{3.78} & \textbf{6.15} & \textbf{2.06} & \textbf{5.20} \\
OLaPh LLM    & 20.25  & 7.71 & 13.31  & 3.20 & 9.02 \\ \bottomrule
\end{tabularx}
\end{table}

\autoref{tab:oov_results} demonstrates that OLaPh maintains its superior performance even on words absent from its lexical vocabulary, validating the efficacy of its language detection and statistical subword segmentation components. This is most evident in the Spanish evaluation; despite having the highest proportion of OOV samples ($30.69\%$), OLaPh's PER on this subset is actually lower than its performance on the full dataset. This indicates that the subword segmentation does not merely function as a fallback mechanism, but a robust and reliable method for G2P conversion.
In contrast, OLaPh exhibits a distinct error increase in English, a trend mirrored by all evaluated frameworks. This is likely attributable to the high frequency of allophonic variations in the language as well a mixups between British and American English. Manual sample checks confirmed that these errors often stem from minor transcriptional variations in the benchmark rather than fundamental failures in phonemic mapping. While a comprehensive normalization of all languages would likely improve PER scores across the board, particularly for English, this was not feasible given the substantial manual effort required for a dataset of this scale.

\subsection{Neural Generalization on OLaPh-Generated Training Data}

Finally, we conducted a sample-wise comparison between the OLaPh framework and the OLaPh LLM. Although the framework achieved a lower overall PER as previously discussed, the LLM outperformed its "teacher" on up to 22.49\% of OOV samples as can be seen in \autoref{tab:framework_llm_comparison}. This suggests that the neural model has developed a degree of phonemic generalization that transcends the deterministic subword segmentation of the OLaPh framework. In these instances, the LLM was able to resolve ambiguities or edge cases where the statistical subword splitting produced sub-optimal results.

\begin{table}[ht]
\centering
\caption{Comparative analysis of OLaPh framework vs. OLaPh LLM on OOV samples.}
\label{tab:framework_llm_comparison}
\begin{tabularx}{\textwidth}{@{} l *{8}{>{\centering\arraybackslash}X} @{}}
\toprule
\textbf{Language} & \textbf{OOV Words} & \textbf{Framework Wins} & \textbf{LLM Wins} & \textbf{Tie} & \textbf{Framework Wins (\%)} & \textbf{LLM Wins (\%)}& \textbf{Ties (\%)} \\ \midrule
EN      & 4,730  & 2,032  &1,064   &1,634   &42.96       &22.49       &34.55\\
FR      & 13,028 & 4,504  &1,323   &7,201   &34.57       &10.16       &55.27\\
DE      & 4,970  & 3,437  &448    &1,085   &69.15       &9.01        &21.83\\
ES      & 30,318 & 3,050  &612    &26,656  &10.06       &2.02 &87.92\\\bottomrule
\end{tabularx}
\end{table}
\section{Discussion}
While the superior performance of the OLaPh framework on the base benchmark evaluation underscores the value of large corpora for reliable lexical lookups, the results on OOV words show that its included backup functionalities act as a solid framework for reliable G2P conversion with high accuracy. Furthermore, OLaPh LLM, despite being trained solely on data synthesized by the framework, demonstrated a "Student-Teacher" breakthrough, outperforming or matching the framework in most languages. This indicates that neural models can generalize phonetic patterns more flexibly than deterministic recursive algorithms, provided they are trained on high-consistency synthetic data. A structured curation process for the training data was not conducted, and only shallow filtering (basic language detection, removal of URLs) was applied to the generally noisy FineWeb corpora, resulting in limited vocabulary diversity in our current training corpus, which constrains the model's generalization capabilities. Expanding and more rigorously selecting training data in future work could close the performance gap between the LLM and framework.

Our results for English highlight the persistent difficulty of G2P for languages with deep orthography. The increased PER across all models as well as our subsequent manual audit suggest that allophonic variation remains a significant noise factor for fully "balanced and fair" evaluations. While further manual normalization of the WikiPron benchmark would likely yield slightly lower PER scores for all evaluated models, our findings suggest that OLaPh performs robust enough to provide high-utility transcriptions when handling linguistic ambiguity. The impact of these minor variations on actual speech quality should be further investigated through TTS downstream evaluations in future work.

The current iteration of OLaPh focuses on four major European languages (English, German, French, and Spanish), yet the framework's modularity allows for rapid expansion through the addition of further lexica. Future research will explore scaling the OLaPh LLM in both parameter count and dataset diversity. By incorporating a larger number of diverse synthetic samples, we anticipate that LLMs with G2P fine-tuning will eventually surpass hybrid frameworks like OLaPh in both OOV generalization and overall accuracy.
\section{Conclusion and outlook}
In this work, we presented OLaPh, a hybrid phonemization framework supporting four major languages. By integrating Named Entity Recognition (NER), POS tagging, and a novel statistical subword segmentation algorithm with a multi-stage backup lookup, OLaPh addresses the core limitations of deterministic G2P systems. A benchmark evaluation has proven its high accuracy and added value over established G2P frameworks and previously published solutions for phonemization based on neural approaches. Furthermore, an instruction-tuned LLM trained on OLaPh-synthesized data demonstrated significant data efficiency, achieving robust generalization for multilingual G2P tasks. This underscores the framework's utility not only as a standalone tool but as a high-fidelity data generator for neural architecture training.

Future work will focus on integrating language models for enhanced context-sensitive disambiguation and expanding the framework to support a broader selection of languages through Wiktionary resources. We also aim to refine the presented fallback mechanisms by transitioning from character-level to token-aware representations, ensuring greater robustness in subword segmentation. Additionally, we will explore scaling the OLaPh LLM in both parameter count and training data diversity. By leveraging larger sets of OLaPh-synthesized data, we anticipate that future iterations of the neural model will eventually surpass the deterministic framework in both OOV generalization and absolute accuracy. Both the OLaPh framework and OLaPh LLM are published for free use at \url{https://github.com/iisys-hof/olaph}.

\bibliographystyle{splncs04}
\bibliography{paper}

\end{document}